\documentclass[10pt,twocolumn,letterpaper]{article}

\usepackage{cvpr}              

\usepackage{textcmds}
\usepackage{adjustbox}
\usepackage{multirow}
\usepackage{amsmath} 
\usepackage{amssymb}  
\usepackage{algorithm}
\usepackage{algpseudocode} 
\usepackage{booktabs} 
\usepackage{siunitx}
\usepackage{float}
\usepackage[table]{xcolor}  
\usepackage{colortbl}
\usepackage{bm}
\usepackage{pifont} 

\usepackage[dvipsnames]{xcolor}
\usepackage{colortbl}

\usepackage{soul}

\definecolor{cvprblue}{rgb}{0.21,0.49,0.74}
\usepackage[pagebackref,breaklinks,colorlinks,allcolors=cvprblue]{hyperref}
\definecolor{lightgray}{gray}{0.85}
\newcolumntype{L}{>{\columncolor{lightgray}}c}
\def\paperID{xxxxx} 
\def\confName{CVPR}
\def\confYear{2026}

\title{Hierarchical Point-Patch Fusion with Adaptive Patch Codebook for 3D Shape Anomaly Detection}

\author{
Xueyang Kang$^{1}$
\and
Zizhao Li$^{1}$
\and
Tian Lan$^{2}$
\and
Dong Gong$^{3}$
\and
Kourosh Khoshelham$^{1}$
\and
Liangliang Nan$^{4}$\thanks{\tt\tiny
Corresponding author: Liangliang.Nan@tudelft.nl}
\\[0.5em]
$^{1}$The University of Melbourne, Parkville, Victoria 3010,   Australia
\\
$^{2}$Tsinghua University, Haidian District, Beijing 100871, China
\\
$^{3}$The University of New South Wales, Sydney, NSW 2052, Australia
\\
$^{4}$Delft University of Technology, Mekelweg 5, 2628 CD Delft, Netherlands
}

\begin{document}
\maketitle
\begin{abstract}
3D shape anomaly detection is a crucial task for industrial inspection and geometric analysis. Existing deep learning approaches typically learn representations of normal shapes and identify anomalies via out-of-distribution feature detection or decoder-based reconstruction. They often fail to generalize across diverse anomaly types and scales, such as global geometric errors (e.g., planar shifts, angle misalignments), and are sensitive to noisy or incomplete local points during training. To address these limitations, we propose a hierarchical point–patch anomaly scoring network that jointly models regional part features and local point features for robust anomaly reasoning. An adaptive patchification module integrates self-supervised decomposition to capture complex structural deviations. Beyond evaluations on public benchmarks (Anomaly-ShapeNet and Real3D-AD), we release an industrial test set with real CAD models exhibiting planar, angular, and structural defects. Experiments on public and industrial datasets show superior AUC-ROC and AUC-PR performance, including over 40\% point-level improvement on the new industrial anomaly type and average object-level gains of 7\% on Real3D-AD and 4\% on Anomaly-ShapeNet, demonstrating strong robustness and generalization. The code and test data are publicly available at: \href{https://github.com/alexandor91/Shape-Anomaly-Codebook.git}{github.com/Shape-Anomaly-Codebook.git}

\end{abstract}    
\section{Introduction}
\label{sec:intro}
\begin{figure}[t]
  \centering
  \includegraphics[width=0.88\columnwidth]{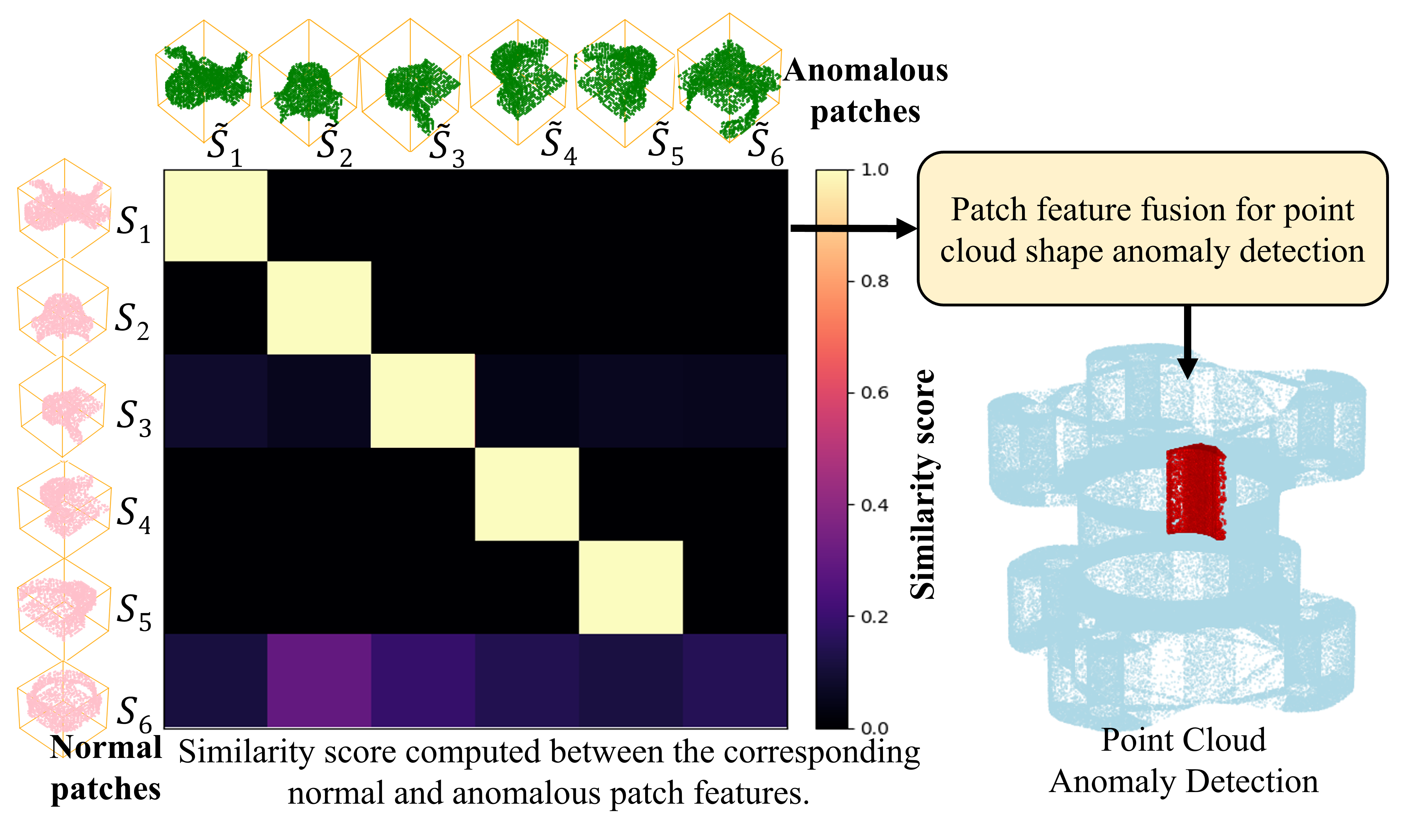}
  \vspace{-1.0em}
    \caption{
    \textbf{Patch feature fusion for 3D shape anomaly detection in point clouds}. 
    Normal patches $\{\mathbf{S}_j\}, j \in \{1,\ldots,6\}$ and abnormal patches $\{\tilde{\mathbf{S}}_j\}$ are ordered by distance to the object center. The left heatmap visualizes the cosine similarity between patch features, where higher values indicate stronger similarity between normal and abnormal patches. Anomalous regions exhibit distinctive feature discrepancies in the last row, indicating ambiguous patch correspondences. These patch-level differences are fused to guide pointwise anomaly detection, with red regions in the right point cloud marking detected anomalies. 
    }
  \label{fig:part-attention}
  \vspace{-1.0em}
\end{figure}

Shape anomaly detection is a fundamental industrial challenge, aiming to identify and localize structural or geometric defects in manufactured 3D products. Conventional autonomous approaches typically rely on geometric comparisons between a normal template and a test shape. A set of normal shapes is encoded into latent features as a reference, and anomalies are detected by comparing the feature distributions using Out-of-Distribution (OOD) detection~\cite{PALM2024, zheng2023out, she23, cider, pei2022out}. Early research focused on 2D image anomaly detection, such as MVTec AD~\cite{bergmann2019mvtec}, where methods detect anomalies by identifying appearance differences (\emph{e.g.}, cracks) between test images and templates~\cite{hu2024anomalydiffusion, li2021cutpaste, liu2023simplenet}. 

With the emergence of 3D anomaly detection benchmarks such as Real3D-AD~\cite{liu2023real3d}, MVTec 3D-AD~\cite{bergmann2021mvtec}, and Anomaly-ShapeNet~\cite{Li_2024_CVPR}, research has increasingly shifted toward 3D geometric domains. Recent approaches, including dual-memory branches~\cite{chu2023shape} and teacher--student architectures~\cite{qin2023teacher}, model the distribution gap between normal and anomalous features. Multimodal methods further improve discrimination by fusing RGB images and point clouds~\cite{long2025revisiting, costanzino2024multimodal, wang20253dkeyad}. Zero-shot and few-shot settings are explored by aligning point clouds with multi-view image projections~\cite{zhou2024pointad, cheng2024towards} or leveraging view synthesis~\cite{zuo2024clip3d}. Despite these advances, most methods struggle with large-scale structural anomalies and planar displacements common in industrial scenarios, as they rely primarily on local point representations. Reconstruction-based methods, such as DRAEM-A~\cite{zavrtanik2021draem} and R3D-AD~\cite{zhou2024r3d}, reconstruct normal shapes and detect anomalies via reconstruction errors, but are sensitive to noise, unstable in planar or concave regions, and depend heavily on negative augmentation. Similarly, point-wise or keypoint-based approaches, including 3DKeyAD~\cite{wang20253dkeyad}, geometric descriptor methods~\cite{bergmann2023anomaly}, and PO3AD~\cite{ye2025po3ad}, focus on per-point regression and are effective mainly for small local defects, while remaining inadequate for large structural displacements or part misalignments that are underrepresented in public benchmarks.

To address these challenges, we first construct a small industrial test dataset consisting of eight real components with anomalies such as planar displacements and large gear-angle misalignments. We then propose an \textbf{adaptive multi-scale patch-based encoding framework} for 3D shape anomaly detection. Our method extracts multi-scale patches from normal shapes and stores their features in a lightweight, position-invariant \textbf{patch feature codebook}, which is queried by pseudo-negative inputs during training to infer patch-level discrepancies. Unlike point-based methods that rely only on local geometry, this design provides multi-scale references for detecting anomalies at different scales, enabling more robust and scale-aware reasoning. Although patch-based decomposition has been explored in segmentation and completion tasks~\cite{alonso2021semi, rao2022patchcomplete}, our approach introduces adaptive patch scales tailored to anomaly detection. During inference, anomalous shapes undergo the same multi-scale patchification, and their patch features are compared with the codebook to compute patch similarity scores. 
To summarize, our main contributions are threefold: 
\begin{itemize}
    \item An \textbf{adaptive patchification module} that extracts multi-scale patches from normal shapes and caches them in a lightweight, position-invariant \textbf{patch codebook}, which captures the invariant geometry of various scales.
    \item \textbf{A patch–point feature fusion model} that integrates patch-level and point-level features via cross-attention. A post-modulation based on patch similarity between normal and anomalous regions enables adaptive discrimination of anomalies across diverse scales and types.
    \item \textbf{A new industrial 3D anomaly test set} featuring realistic manufacturing defects such as gear cracks and planar displacements. Extensive experiments demonstrate superior performance with over \textbf{40\%} improvement on industrial anomalies and average gains of \textbf{7\%} on Real3D-AD, highlighting strong robustness over state-of-the-art baselines.
\end{itemize}





\section{Related Work}
\label{sec:formatting}

\noindent\textbf{Anomaly detection applications.}
The MVTec-AD~\cite{bergmann2019mvtec} benchmark catalyzed extensive research in image anomaly detection, spawning methods that model normal distributions or synthesize pseudo anomalies~\cite{zheng2023out, hu2024anomalydiffusion, liu2023simplenet, schluter2022natural, zavrtanik2021draem, zhang2024realnet}. In 3D vision, shape anomaly detection identifies geometric deviations from learned normal shapes through \emph{OOD detection}, where anomalous inputs fall outside the normal feature distribution. MVTec-3D-AD~\cite{bergmann2021mvtec} enabled multi-modal methods combining RGB and point cloud~\cite{wang2025m3dm, zuo2024clip3d, costanzino2024multimodal, wang2023multimodal}, though gains over single-modality approaches remain marginal. OOD detection extends beyond industrial inspection to medical imaging~\cite{zimmerer2022mood, heer2021ood} and hyperspectral sensing~\cite{yang2023full, zimmerer2022mood}, with summary surveys available~\cite{li2025out}.

\noindent\textbf{Reconstruction-based anomaly detection.}
A major line of anomaly detection research relies on reconstruction, where models learn to reproduce normal inputs and detect deviations~\cite{zavrtanik2021draem}. These methods train models to learn geometric representations of normal shapes and reconstruct point clouds for comparison with inputs at inference. IMRNet~\cite{Li_2024_CVPR}, for instance, employs a masked autoencoder to reconstruct point clouds, while R3D-AD~\cite{zhou2024r3d} leverages diffusion to recover normal distributions. Lastly, PO3AD~\cite{ye2025po3ad} directly regresses point-wise offsets between pseudo anomalies and normal inputs. 

\noindent\textbf{Memory bank-based anomaly detection.}
Memory bank mechanisms have been extensively adopted in computer vision tasks such as semantic segmentation~\cite{alonso2021semi, zhou2024rmem} and object detection~\cite{sun2021mamba}. For anomaly detection, early works integrated a memory module with an autoencoder to store normal patterns~\cite{gong2019memorizing}, retrieving the closest memory feature for reconstruction, where large reconstruction errors signal anomalies. Subsequent methods extended this concept to geometric data by mapping local shape features into a memory space representing the normal feature distribution, so that anomalous inputs are naturally projected out-of-distribution. Some approaches employ hybrid representation learning across RGB and point cloud modalities~\cite{zhou2024pointad} for anomaly detection. In contrast, others use a shared cross-modal memory~\cite{wang2023multimodal} to fuse visual and geometric features before anomaly classification.

\section{Method}
We generate pseudo-abnormal shapes $\tilde{\mathbf{S}}$ from normal inputs $\mathbf{S}$ via negative augmentation~\cite{ye2025po3ad}, enabling robust anomaly discrimination. During inference, only $\tilde{\mathbf{S}}$ is provided, with patch features retrieved from a pre-cached normal codebook. Both $\mathbf{S} \in \mathbb{R}^{N \times 3}$ and $\tilde{\mathbf{S}} \in \mathbb{R}^{N \times 3}$ are divided into patches ${\mathbf{S}_j}$ and ${\tilde{\mathbf{S}}_j}$ via \textbf{Farthest Point Sampling (FPS) patchification}. The \textbf{patch feature extraction module (a)} encodes patch centers into normal features $\mathbf{t}_k \in \mathbb{R}^{d}$ (template) and anomalous features $\mathbf{p}_j \in \mathbb{R}^{d}$. Normal features populate the \textbf{shape patch codebook (b)} for anomaly retrieval. During inference, patch correspondences between $\mathbf{p}_j$ and $\mathbf{t}_k$ are established via similarity-based querying, followed by \textbf{point-to-patch cross-attention (c)} that fuses local point features with nearest normal patch embeddings. The \textbf{part score modulation module (d)} refines fused features through gated modulation and predicts point-wise offset $\hat{\mathbf{o}} \in \mathbb{R}^{N \times 3}$, which is mapped by scoring function $\phi: \mathbb{R}^{N \times 3} \rightarrow \mathbb{R}$ to normalized anomaly scores in $[0,1]$. 
\begin{figure*}[!th]
\centering
  \includegraphics[width=0.94\linewidth]{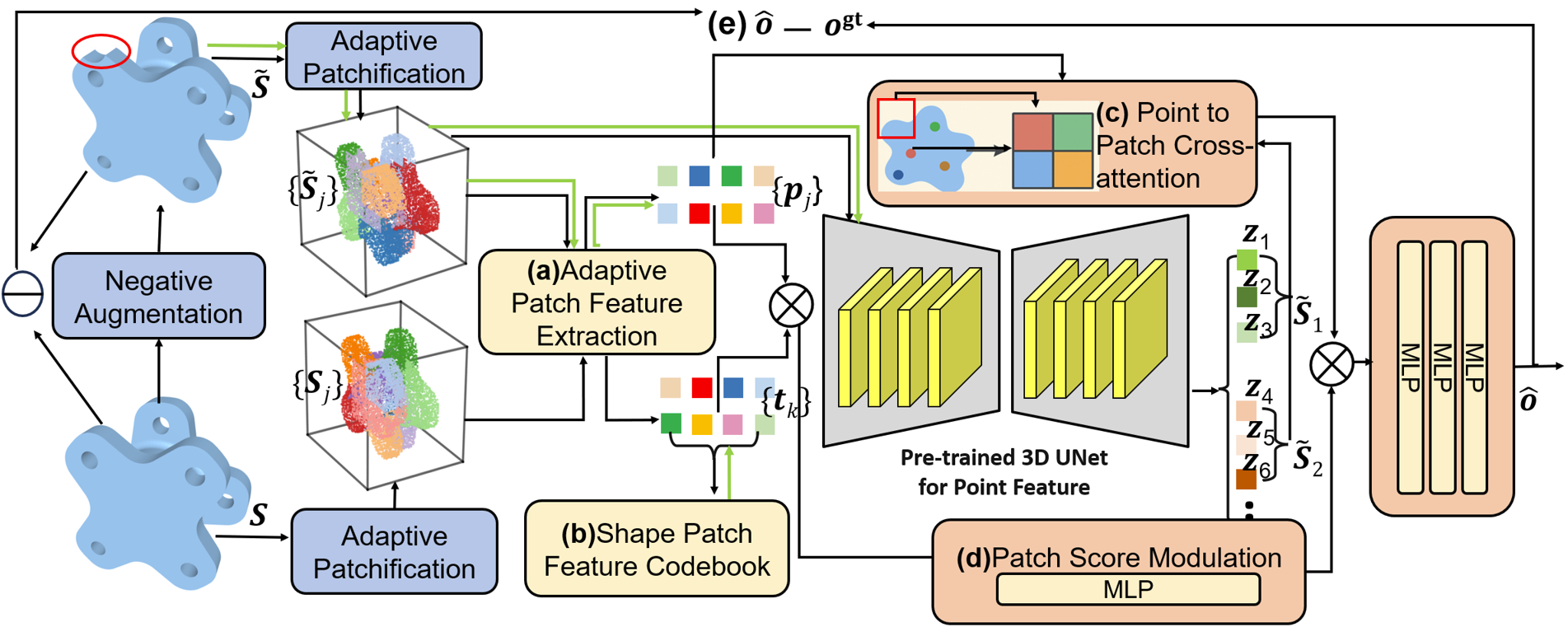}
  \vspace{-0.8em}
 \caption{\textbf{Overview of the proposed shape anomaly detection framework}. The pipeline processes a normal shape $\mathbf{S}$ through several modules: \textbf{Patchification} segments the input into patches $\{\mathbf{S}_j\}$, while \textbf{Negative Augmentation} generates pseudo anomalous shapes $\tilde{\mathbf{S}}$ that are also patchified during training (red regions indicate anomalies); (a) \textbf{Adaptive Patch Feature Extraction} computes features for each patch by averaging its points and querying a pre-trained UNet to obtain patch features $\{\mathbf{p}_j\}$; (b) \textbf{Template Codebook} stores a dictionary of normal template patch features $\{\mathbf{t}_k\}$, from which each patch retrieves its closest matching patch; (c) \textbf{Point to Patch Cross-attention} fuses the template features $\{\mathbf{t}_k\}$ with point-level features extracted from the pre-trained UNet encoder through multi-head attention mechanism; (d) \textbf{Patch Score Modulation} modulates the cross-attention outputs using similarity scores between normal and abnormal patch features $\{\mathbf{t}_k\}, \{\mathbf{p}_j\}$, with the modulated features fed into MLP regression layers to predict the anomaly score $\delta_{\text{pred}}$. During training, (e) the loss is computed as $\hat{\mathbf{o}} - \mathbf{o}_{gt}$, where $\mathbf{o}_{gt}$ represents the ground-truth offset direction derived from the difference $\mathbf{S} - \tilde{\mathbf{S}}$. During inference (indicated by green arrows), only the test shape is input for anomaly detection.}
  \label{fig:overview}
\vspace{-0.6em}
\end{figure*}

\subsection{Patchification \& negative augmentation} During training, given a normal shape $\mathbf{S}$, we apply \textbf{negative augmentation} to generate pseudo-anomalous shapes $\tilde{\mathbf{S}}$. Specifically, random points on $\mathbf{S}$ are selected as anomaly centers, around which local geometric deformations are introduced using Gaussian kernel-based displacements guided by the local normal directions. Positive and negative displacements simulate concave or sink-type anomalies, following the strategy of Real3D-AD \cite{liu2023real3d}. To further simulate bulging deformations, the normal directions are modulated by a sine wave function, creating alternating signs of deformation. Additionally, random holes and planar cut-offs using cubic or cylindrical masks are applied to mimic missing or displaced surface regions. This augmentation strategy generates diverse, realistic structural defects that resemble those observed in real test cases.  

We then adopt an \textbf{adaptive patchification} process that extracts patches at three scales (\emph{e.g., {8, 32, 64}}), providing multi-scale references that help the model distinguish anomalies of varying sizes and improve robustness to noisy points. Specifically, for each level $l \in \{1,2,3\}$, patches of size $p_l$ are sampled from $\mathbf{S}$, where $\mathbf{p}^1 < \mathbf{p}^2 < \mathbf{p}^3$ correspond to fine, medium, and coarse resolutions, respectively. These multi-scale patches capture geometric details across varying granularities, enabling the network to model local fine structures and broader contextual cues jointly.


\subsection{Adaptive patch feature extraction}
For each patch at level $l \in \{1,2,3\}$, denoted as $\tilde{\mathbf{S}}_j^{(l)}$ containing points 
$\{\mathbf{x}_i^{j,(l)} \in \mathbb{R}^3 \mid i \in \mathcal{N}_j^{(l)}\}$, where $\mathcal{N}_j^{(l)}$ denotes the point set of patch $\tilde{\mathbf{S}}_j^{(l)}$, point-wise features are extracted using a pre-trained 3D UNet encoder. Each point $\mathbf{x}_i^{j,(l)}$ is processed to obtain its feature representation $\mathbf{z}_i^{(l)}$. The corresponding patch feature $\mathbf{p}_j^{(l)}$ is computed as the mean of all point coordinates relative to the patch centroid and processed by the pre-trained encoder $f$:
\begin{equation}    
\mathbf{p}_j^{(l)} = f\Bigg(\frac{1}{|\mathcal{N}_j^{(l)}|} \sum_{i \in \mathcal{N}_j^{(l)}} (\mathbf{x}_i^{j,(l)} - \mathbf{c}_j^{(l)})\Bigg).
\label{eq:patch-center-feature}
\end{equation}
Note that the patch center $\Bar{\mathbf{x}}_i^{(l)}$ for the query differs from the geometric sphere cluster center $\mathbf{c}_j^{(l)}$ used for encoder querying. Specifically, $\mathbf{p}_j^{(l)}$ denotes the centroid feature computed from the actual point distribution within the patch, whereas the center feature corresponds to the geometric sphere center. For clarity, we omit the superscript $j$ and level $l$ in subsequent notations; all point indices $i$ implicitly correspond to their associated patch $j$ and scale level $l$. The same process applies to template patch feature $\mathbf{t}_k^{()}$.

\subsection{Shape patch feature codebook}
During training, the extracted patch features at scale $l$, 
$\{\mathbf{t}_k^{(l)}\}, k \in \mathcal{N}_{k}^{(l)}$, from the normal shape input $\mathbf{S}$ are stored in a multi-scale \textbf{shape patch feature codebook}, where the hash key of each patch position serves as the key, and the corresponding patch features form the value set. The patch features are ordered by their distance from the patch center to the object centroid. To reduce redundancy and limit memory usage, the codebook patch features are extracted from the relative point distribution within each patch, ensuring translation invariance. When symmetry is present, as shown in the input of~\cref{fig:overview}, identical corner regions share the same patch feature, which is stored once and referenced by position for reuse during repeated retrieval. This process is repeated independently for each scale $l \in \{1,2,3\}$ to build codebooks for fine, medium, and coarse resolutions.

During inference, the patch features of a test input at scale $l$, $\{\mathbf{p}_j^{(l)}\}$, serve as queries to retrieve the most similar entries $\{\mathbf{t}_j^{(l)}\}$ from the corresponding scale codebook using cosine similarity. The overall similarity score at each scale is computed by summing the feature similarities along the diagonal positions:
\begin{equation}
\mathcal{\alpha}^{(l)} = \sum_{j=1}^{\mathcal{N}_{\text{Patch}}^{(l)}} 
(\mathbf{p}_j^{(l)} \cdot \mathbf{t}_j^{(l)}),
\label{eq:patch-similarity}
\end{equation}
\noindent where a higher $\mathcal{\alpha}^{(l)}$ indicates stronger correspondence between the test and normal patches at scale $l$. The scale $l^*$ with the maximal similarity sum is selected through $\mathrm{argmax}_{l \in {1,2,3}} \mathcal{\alpha}^{(l)}$, and the retrieved patch features at scale $l^*$ are used to guide anomaly-aware feature modulation for final prediction.  
This multi-scale retrieval ensures that both local and global patch correspondences are considered, enhancing robustness to geometric scale variations.

The codebook update follows a thresholded similarity scheme. Each normalized patch feature is compared to existing entries using cosine similarity; features exceeding a threshold $\tau$ are merged via count-weighted averaging to suppress redundancy, while dissimilar features create new entries to preserve geometric diversity. Spatial hash keys from 3D positions are stored to maintain coverage across locations. We provide the pseudo code for the algorithm implementation in \cref{alg:codebook-update}.
\vspace{-0.8em}
\begin{algorithm}[!th]
\centering
\caption{Adaptive Patch Codebook Update}
\label{alg:codebook-update}
\begin{algorithmic}[1]
\setlength{\itemsep}{-1.32pt}
\setlength{\parskip}{-1.32pt}
\Require Patch feature $\mathbf{t}_j^{(l)} \in \mathbb{R}^{32}$, hash keys $\mathcal{H}_j^{(l)} = H^{(l)}(\mathbf{p}_j = (x_j, y_j, z_j))$, codebook $\mathcal{C}^{(l)}$, threshold $\tau$, Merging times $n_i$
\Ensure Updated codebook $\mathcal{C}^{(l)}$
\State Normalize: $\mathbf{t}_j^{(l)} \leftarrow \mathbf{t}_j^{(l)} / \|\mathbf{t}_j^{(l)}\|_2$
\For{each $(\mathbf{c}_i^{(l)}, \mathcal{H}_i^{(l)}) \in \mathcal{C}^{(l)}$} \Comment{Iterate over codebook}
    \State $s_i \leftarrow (\mathbf{t}_j^{(l)})^\top \mathbf{c}_i^{(l)}$ \Comment{Cosine similarity}
    \If{$s_i \geq (\tau = 0.85)$} \Comment{Merge similar features}
    \State $\mathbf{c}_i^{(l)} \leftarrow \frac{n_i \mathbf{c}_i^{(l)} + s_i \mathbf{t}_j^{(l)}}{n_i + s_i}$, \Comment{Count-weighted merge}        
        \State $\mathcal{H}_i^{(l)} \leftarrow \mathcal{H}_i^{(l)} \cup \{H^{(l)}(\mathbf{p}_j)\}$ \Comment{Add position hash}
        \State \Return $\mathcal{C}^{(l)}$
    \EndIf
\EndFor
\State $\mathcal{C}^{(l)} \leftarrow \mathcal{C}^{(l)} \cup \{(\mathbf{t}_j^{(l)}, \{H^{(l)}(\mathbf{p}_j)\})\}$ \Comment{Add new entry}
\State \Return $\mathcal{C}^{(l)}$
\end{algorithmic}
\end{algorithm}


    


\subsection{Point-to-patch cross attention}
Each point coordinate $\mathbf{x}_i$ from the pre-trained U-Net encoder is associated with a point feature $\mathbf{z}_i$. 
During inference, the point features $\{\mathbf{z}_i\}$ are mapped to query tokens $\{\mathbf{q}_i\}$, while the retrieved patch codebook features $\{\mathbf{t}_k\}$ serve as key-value tokens $\{\mathbf{k}_i, \mathbf{v}_i\}$ in a RoPE-based multi-head cross-attention module.

\paragraph{Rotary position embedding.} 
To incorporate relative spatial relationships between points and their corresponding patches, we adopt Rotary Position Embedding (RoPE) \cite{su2024roformer}. RoPE encodes relative positional information by rotating feature embedding according to the angular offset between a point $\mathbf{x}_i$ and its patch center $\Bar{\mathbf{x}}i$: 
\begin{equation}
\label{eq:rope}
\text{RoPE}(\mathbf{x}_i, \Bar{\mathbf{x}}_i) = \mathbf{R}_{\Theta,i}^{d}\,(\mathbf{x}_i - \Bar{\mathbf{x}}_i ).
\end{equation}
where $\mathbf{R}_{\Theta, i}^{d}$ is a rotation matrix applied over the $d$-D embedding space and parameterized by frequency $\Theta$.

\paragraph{RoPE-enhanced cross attention.}
Given the RoPE-transformed embeddings, cross-attention between point and patch tokens is then computed as,
\begin{align}
    \label{eq:rope-attn}
\text{Attn}(\mathbf{q},\mathbf{k},\mathbf{v}) 
= \sum_{i=1}^{N} 
\alpha_i \, \mathbf{v}_i = \hat{\mathbf{z}}_i,\\
\alpha_i = \frac{ \sum_{n=1}^{N} \left( \mathbf{R}_{\Theta,n}^{d} \, \phi(\mathbf{q}_n) \right)^{\!\top} \left( \mathbf{R}_{\Theta,n}^{d} \, \varphi(\mathbf{k}_n) \right) \mathbf{v}_n }{ \sum_{n=1}^{N} \phi(\mathbf{q}_n)^{\!\top} \varphi(\mathbf{k}_n) },
\end{align}
\noindent where $\mathbf{q}_n$, $\mathbf{k}_n$, and $\mathbf{v}_n$ denote the query, key, and value, respectively.  $\phi(\cdot)$ and  $\varphi(\cdot)$ are non-negative feature mappings, typically $\phi(x) = \varphi(x) = \mathrm{elu}(x) + 1$. Multi-head concatenation followed by linear projection yields updated point tokens $\{\hat{\mathbf{z}}_i\}$, enabling patch-guided, geometry-aware feature fusion for anomaly detection. This cross-attention mechanism embeds patch-level priors into point-level features. By attending to retrieved normal patch embeddings, each point aligns with its normal patch geometry, improving detection of both fine-grained local defects and large structural misalignments.
\subsection{Patch score modulation}
\noindent
Given the normalized patch features $\{\mathbf{p}_j\}$ of an anomalous shape and the patch features $\{\mathbf{t}_k\}$ of a normal shape, the patch-level discrepancy is first computed as $\Delta \mathbf{f}_{kj} =  (1-\mathbf{t}_k \cdot \mathbf{p}_j)$ between the matched normal patch $k$ and the corresponding anomaly patch $j$. A gating function $\sigma_i(\cdot)$ estimates the anomaly likelihood $\mathbf{\rho}_{i} = \sigma_i(\text{MLP}_{\text{gate}}(\Delta \mathbf{f}_{kj}))$ of $i$th feature, while a modulation network $\mathcal{M}_{\text{mod}}(\cdot)$ predicts the feature-wise scaling and shifting parameters $(\boldsymbol{\gamma}_i, \boldsymbol{\beta}_i) = \mathcal{M}_{\text{mod}}(\Delta \mathbf{f}_{kj})$. The final modulated feature is defined as: 
\begin{equation}
\label{eq:gated_modulation}
\mathbf{z}'_i = 
\mathbf{\rho}_{i} \odot 
\left(
\boldsymbol{\gamma}_i \odot  \hat{\mathbf{z}}_i
+ \boldsymbol{\beta}_i
\right),
\end{equation}
where $\hat{\mathbf{z}}_i$ is the cross-attended feature in~\cref{eq:rope-attn}, $\odot$ represents element-wise multiplication, and $\mathbf{z}'_{i}$ is the modulated point feature corresponding to the $i$-th point within the $j$-th patch after RoPE-enhanced cross-attention. The gate $\mathbf{\rho}_{i} \in [0,1]$ adaptively weights each patch contribution to the anomaly score based on the inverse feature similarity in~\cref{eq:patch-similarity}. 

Finally, the refined point features $\mathbf{z}'_i$ are concatenated with the attended point features $\hat{\mathbf{z}}_i$ and passed through a multilayer perceptron (MLP) for residual prediction of the anomaly offset vector $\hat{\mathbf{o}}_i$:
\begin{equation}
\hat{\mathbf{o}}_i = 
\text{MLP}\left({Concat}[\mathbf{z}'_i,\, \hat{\mathbf{z}}_i]\right) + \hat{\mathbf{z}}_i,
\end{equation}
where $\hat{\mathbf{o}}_i$ represents the predicted anomaly offset vector for each point. This gated modulation mechanism adaptively emphasizes discriminative patch–point relationships to detect precise shape anomalies of varying sizes.

\subsection{Loss function}
\noindent
For each point $\mathbf{x}_i$ in an abnormal shape, let $\mathbf{o}_i^{\mathrm{gt}}\!\in\!\mathbb{R}^3$ denote the ground-truth offset vector from the abnormal point to its corresponding normal counterpart, and $\hat{\mathbf{o}}_i$ be the predicted offset. The network jointly predicts (a) the point-wise anomaly offset $\hat{\mathbf{o}}_i$, and (b) a direction mask $\hat{m}_i$ indicating the sign of deviation. The overall anomaly loss combines three complementary terms:
\begin{equation}
\mathcal{L}_{\text{anomaly}}
= \mathcal{L}_{\text{dist}}
+ \lambda_{\text{sim}}\mathcal{L}_{\text{sim}}
+ \lambda_{\text{BCE}}\mathcal{L}_{\text{BCE}},
\label{eq:combination-loss}
\end{equation}
where $\lambda_{\text{sim}}$ and $\lambda_{\text{BCE}}$ balance the offset direction and mask label supervision.
The offset regression loss minimizes the mean absolute distance between predicted and ground-truth offset vectors:
\begin{equation}
\mathcal{L}_{\text{dist}}
= \frac{1}{N}\sum_{i=1}^{N} 
\big\| \hat{\mathbf{o}}_i - \mathbf{o}_i^{\mathrm{gt}} \big\|_1.
\end{equation}

\noindent
To encourage alignment between the predicted and true offset directions, a cosine similarity loss is used:
\begin{equation}
\mathcal{L}_{\text{sim}}
= -\frac{1}{N}\sum_{i=1}^{N}
\frac{1}{2}\!\left(
1 + 
\frac{
\hat{\mathbf{o}}_i \cdot \mathbf{o}_i^{\mathrm{gt}}
}{
\|\hat{\mathbf{o}}_i\|_2 \, \|\mathbf{o}_i^{\mathrm{gt}}\|_2 + \epsilon
}
\right),
\end{equation}
where $\epsilon$ is a small constant ($10^{-6}$) for numerical stability. Finally, a binary cross-entropy loss supervises the predicted anomaly sign mask $\hat{m}_i$ against its ground-truth label $m_i^{\mathrm{gt}}$:
\begin{equation}
\mathcal{L}_{\text{BCE}}
= -\frac{1}{N}\sum_{i=1}^{N}
\big[
m_i^{\mathrm{gt}}\log(\hat{m}_i)
+ (1 - m_i^{\mathrm{gt}})\log(1 - \hat{m}_i)
\big].
\end{equation}

\noindent
This combined loss jointly constrains the magnitude, direction, and offset sign consistency of anomaly predictions, enabling stable training and accurate point-wise shape anomaly localization. During inference, the predicted offset magnitude $\hat{\mathbf{o}}_i$ is normalized into the anomaly score $\hat{\delta}_i$ using the same $\mathcal{L}_1$ normalization strategy as in PO3AD~\cite{ye2025po3ad}, yielding anomaly score $\hat{\delta}_i$ within $[0,1]$. Final anomaly scores are calculated solely for points designated as valid by the mask.
\section{Experiments \& Results}
\subsection{Experimental setup}
\paragraph{Datasets \& preprocessing.} 
We primarily use two benchmark 3D point-cloud anomaly-detection datasets for evaluation, including \textbf{Real3D-AD}~\cite{liu2023real3d} and \textbf{Anomaly ShapeNet}~\cite{Li_2024_CVPR}. 
The Anomaly ShapeNet dataset contains 40 categories with a total of 1,600 samples, including both normal and anomalous shapes. 
Real3D-AD includes 12 categories, each with 4 normal template samples and 100 test samples containing anomalies. 
Additionally, we introduce \textbf{a new industrial evaluation dataset} comprising 8 normal samples and 8 corresponding test samples with angular misalignment and planar displacements. 
Since Real3D-AD test scans are half-sided, while the training data are full ${360}^{\circ}$ scans, we adopt the same half-sided configuration during codebook construction for consistency.

For preprocessing, negative augmentation is applied to two geometric scales ($0.01$ and $0.1$), while angular displacement uses intersection cut-off ratios between $0.1$ and $0.5$. 
All inputs are normalized into canonical space and voxelized into ${256}^3$ grids. 
For Anomaly ShapeNet and Real3D-AD, multi-scale patching is performed at three scales with patch numbers ${32, 64, 192}$ and corresponding patch sizes ${64, 32, 8}$. 
For our own test set, we adopt inverse patch scaling with patch numbers ${8, 32, 64}$ and corresponding patch sizes ${192, 64, 32}$ to emphasize large structural variations.

\paragraph{Baselines.} 
We follow recent state-of-the-art learning-based baselines, including PO3AD~\cite{ye2025po3ad}, R3D-AD~\cite{zhou2024r3d}, and ISMP~\cite{ismp_2025}, using their official implementations and default configurations. 
For qualitative comparison, we visualize the results using the provided training code for these methods. 
Since pre-trained weights are unavailable for several categories, only PO3AD is evaluated on our newly created test samples, as other models fail to generalize to unseen categories. 
Evaluation metrics include the \textit{Area Under the Receiver Operating Characteristic Curve (AUC-ROC)} and the \textit{Area Under the Precision-Recall Curve (AUC-PR)}.

\paragraph{Training \& evaluation.} 
All categories are trained for 1500 epochs using the Adam optimizer with a learning rate of $1 \times 10^{-3}$. The weighting coefficients ${\lambda}_{\text{sim}}$ and ${\lambda}_{\text{BCE}}$ in~\cref{eq:combination-loss} are both set to $0.5$. 
The point feature extraction backbone is a Minkowski 3D U-Net~\cite{choy20194d} pre-trained on large-scale 3D data. For our own test samples, $10{,}000$ points are uniformly sampled from each mesh, covering both normal and anomalous cases for evaluation.

\subsection{Evaluation results }
We present quantitative results of object-level AUC-ROC performance across three 3D anomaly detection datasets. Specifically, results on Anomaly-ShapeNet are shown in~\cref{tab:cmps-anomaly-shapenet} with anomaly portions from 1\% to 7\%, results on Real3D-AD in~\cref{tab:cmps-real3dad} with portions from 1\% to 5.4\%, and results on our Industry dataset in~\cref{tab:industry-cmps} where the anomaly ratio varies from below 1\% to over 80\%. Since each class in the proposed test set contains only one instance, object-level metrics are not applicable; thus, we use point-level metrics. For point-level results on the other two datasets, please refer to the supplementary material. These datasets collectively cover a broad range of anomaly densities and spatial distributions.

As shown in~\cref {tab:cmps-anomaly-shapenet}, our approach achieves state-of-the-art average performance across 40 classes. Although it is slightly lower (by only 3\%–5\%) than PO3AD~\cite{ye2025po3ad} on a few cases such as ``bottle0'', ``eraser0'', and ``helmet0'', the overall results remain dominant. Note that class-specific weights are unavailable in PO3AD, preventing exact reproduction; thus, we report the paper's original results. As shown in~\cref{tab:cmps-real3dad}, although a few classes (e.g., ``Chicken'', ``Diamond'', and ``Fish'') have the second-best method outperforming our method, our method still averages a 7.5\% margin over the second-best method. 

On our Industry dataset (\cref{tab:industry-cmps}) with challenging angular misalignment and planar displacements, our model substantially outperforms all baselines. R3D-AD~\cite{zhou2024r3d} fails on most samples, while our model exceeds PO3AD~\cite{ye2025po3ad} by over 50\% in point-level AUC-ROC, demonstrating strong robustness under real industrial conditions. We further provide Point-level AUC-ROC($\uparrow$) results for Anomaly-ShapeNet and Real3DAD in the supplementary file.

Qualitative comparisons in~\cref{fig:baseline-cmps} cover diverse anomaly types including angular shifts, bending, concavity, surface sinks, and bulges. While R3D-AD~\cite{zhou2024r3d} produces unstable detections and PO3AD~\cite{ye2025po3ad} yields many false positives, our method consistently delivers accurate localization aligned with ground truth across all datasets. 

\begin{figure*}[!th]
  \includegraphics[width=\textwidth]{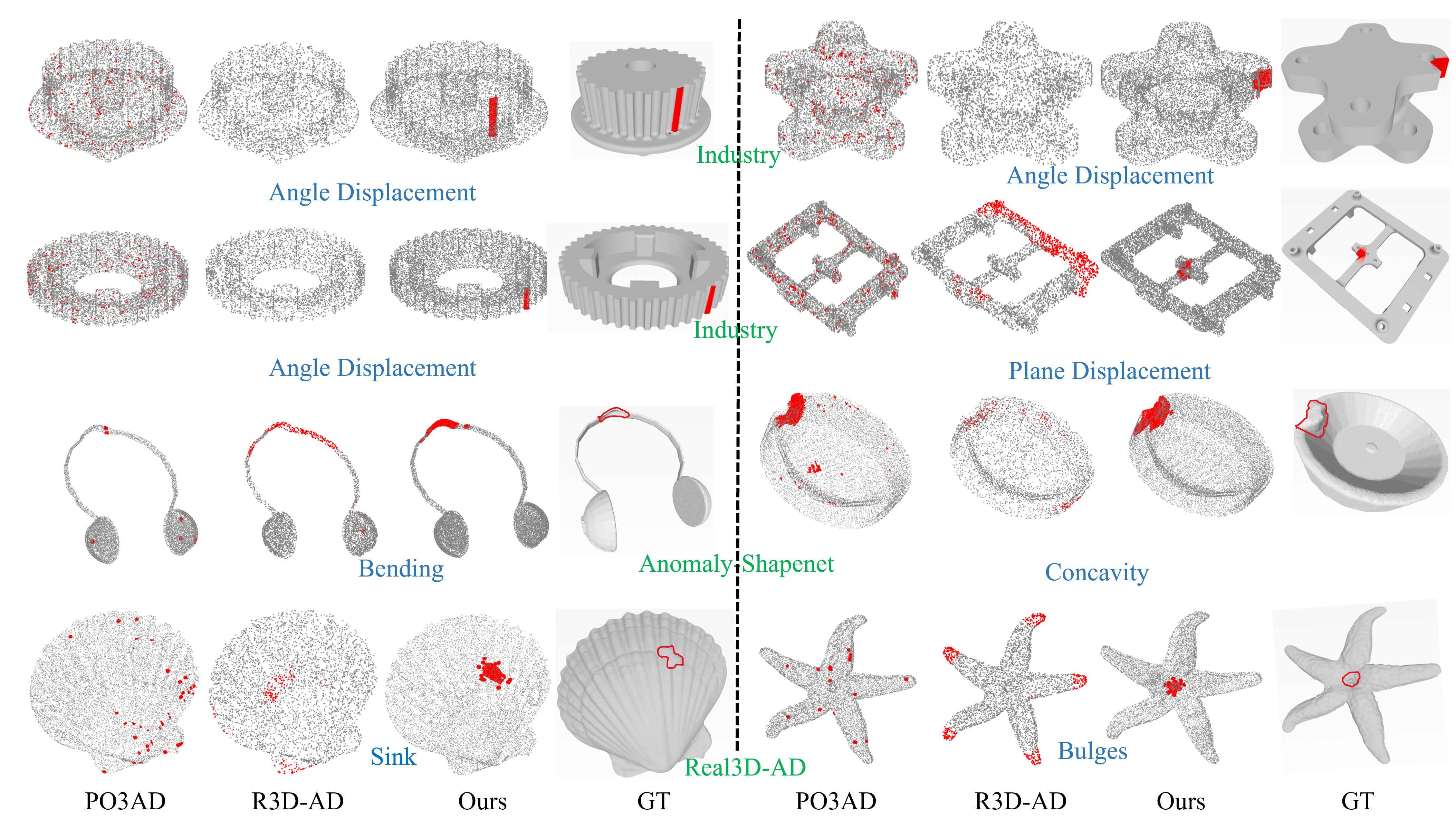}
  \vspace{-2.2em}
\caption{Qualitative comparison on three datasets (\textcolor{green!40!black}{green letters}): our crafted industrial components (1st–2nd rows), Real3D-AD (3rd), and Anomaly-ShapeNet (4th). The dashed line indicates different object classes. Representative anomaly types (\textcolor{blue!80!black}{in blue}) include displacement, bending, concavity, sinks, and bulges. Anomaly points are highlighted in red. Ground truth uses red masks (first two rows) or red contours (last two rows) to indicate anomaly regions overlaid with an anomalous mesh. Each column shows results from a specific method.}
  \label{fig:baseline-cmps}
\end{figure*}

\begin{table*}[!t]
\centering
\caption{Comparison of object-level AUC-ROC results (\%) of various methods on the Anomaly-ShapeNet dataset.}
\vspace{-0.8em}
\label{tab:cmps-anomaly-shapenet}
\resizebox{\linewidth}{!}{
\begin{tabular}{l|cccccccccccccc}
\toprule
Method & ashtray0 & bag0 & bottle0 & bottle1 & bottle3 & bowl0 & bowl1 & bowl2 & bowl3 & bowl4 & bowl5 & bucket0 & bucket1 & cap0 \\ 
\midrule
CPMF \cite{cao2024complementary} & 35.3 & 64.3 & 52.0 & 48.2 & 40.5 & 78.3 & 63.9 & 62.5 & 65.8 & 68.3 & 68.5 & 48.2 & 60.1 & 60.1\\
Reg3D-AD (\cite{liu2023real3d}) & 59.7 & 70.6 & 48.6 & 69.5 & 52.5 & 67.1 & 52.5 & 49.0 & 34.8 & 66.3 & 59.3 & 61.0 & 75.2 & 69.3\\
IMRNet \cite{Li_2024_CVPR} & 67.1 & 66.0 & 55.2 & 70.0 & 64.0 & 68.1 & 70.2 & 68.5 & 59.9 & 67.6 & {71.0} & 58.0 & { 77.1} & 73.7 \\
R3D-AD \cite{zhou2024r3d} & {83.3} & {72.0} & {73.3}    & {73.7} & {78.1} & {81.9} & {77.8} & {74.1} & {76.7} & {74.4} & 65.6 & {68.3} & 75.6 & {82.2}\\
ISMP \cite{ismp_2025} & 86.5 & 73.4 & 72.2 & 86.9 & 74.0 & 76.2 & 70.2 & 70.6 & 85.1 & 75.3 & 73.3 & 54.5 & 68.3 & 67.2 \\
PO3AD \cite{ye2025po3ad} & {100.0} & {83.3} & \textbf{90.0} & 93.3 & \textbf{92.6} & 92.2 & {82.9} & {83.3} & \textbf{88.1} & \textbf{98.1} & {84.9} & {85.3} & {78.7} & {87.7} \\
\rowcolor{lightgray} Ours & \textbf{100.0} & \textbf{85.2} & 87.9 & \textbf{93.8} & {91.3} & \textbf{94.8} & \textbf{88.4} & \textbf{86.7} & {86.9} & {97.1} & \textbf{89.3} & \textbf{87.0} & \textbf{83.9} & \textbf{90.5}\\

\bottomrule
\end{tabular}
}
\resizebox{\linewidth}{!}{
\begin{tabular}{l|cccccccccccccc}
\toprule
Method & cap3 & cap4 & cap5 & cup0 & cup1 & eraser0  & headset0 & headset1 & helmet0 & helmet1 & helmet2 & helmet3  & jar0 & micro. \\ \midrule
CPMF \cite{cao2024complementary} & 55.1 & 55.3 & { 69.7} & 49.7 & 49.9 & 68.9 & 64.3 & 45.8 & 55.5 & 58.9 & 46.2 & 52.0 & 61.0 & 50.9\\
Reg3D-AD \cite{liu2023real3d} & 72.5 & 64.3 & 46.7 & 51.0 & 53.8 & 34.3 & 53.7 & 61.0 & 60.0 & 38.1 & 61.4 & 36.7 & 59.2 & 41.4\\
IMRNet \cite{Li_2024_CVPR} & { 77.5} & 65.2 & 65.2 & 64.3 & { 75.7} & 54.8 & 72.0 & 67.6 & 59.7 & 60.0 & { 64.1} & 57.3 & 78.0 & 75.5\\
R3D-AD \cite{zhou2024r3d} & 73.0 & 68.1 & 67.0 & { 77.6} & { 75.7} & { 89.0} & { 73.8} & { 79.5} & { 75.7} & { 72.0} & 63.3 & { 70.7} & { 83.8} & { 76.2}\\
ISMP \cite{ismp_2025} & 77.5 & 66.1 & 77.0 & 55.2 & 85.1 & 52.4 & 47.2 & 84.3 & 61.5 & 60.3 & 56.8 & 52.2 & 66.1 & 60.0
 \\
PO3AD \cite{ye2025po3ad} & {85.9} & {79.2} & 67.0 & {87.1} & {83.3} & \textbf{99.5} & 80.8 & {92.3} & 76.2 & \textbf{96.1} & {86.9} & {75.4} & {86.6} & {77.6} \\ 
\rowcolor{lightgray} Ours & \textbf{86.4} & \textbf{79.8} & \textbf{80.1} & \textbf{89.2} & \textbf{85.6} & {94.0} & \textbf{81.7} & \textbf{92.8} & \textbf{78.1} & 93.7 & \textbf{88.4} & \textbf{83.9} & \textbf{88.7} & \textbf{80.3}\\

\bottomrule
\end{tabular}
}
\resizebox{\linewidth}{!}{
\begin{tabular}{l@{\hspace{-1pt}}|cccccccccccc|cc}
\toprule
Method & shelf0 & tap0 & tap1 & vase0 & vase1 & vase2 & vase3 & vase4 & vase5 & vase7 & vase8 & vase9 & Average &\\ \midrule
CPMF \cite{cao2024complementary} & 68.5 & 35.9 & 69.7 & 45.1 & 34.5 & 58.2 & 58.2 & 51.4 & 61.8 & 39.7 & 52.9 & 60.9 & 55.9 & \\
Reg3D-AD \cite{liu2023real3d} & { 68.8} & 67.6 & 64.1 & 53.3 & 70.2 & 60.5 & 65.0 & 50.0 & 52.0 & 46.2 & 62.0 & 59.4 & 57.2 & \\
IMRNet \cite{Li_2024_CVPR} & 60.3 & 67.6 & 69.6 & 53.3 &{75.7} & 61.4 & 70.0 & 52.4 & 67.6 & 63.5 & 63.0 & 59.4 & 66.1 &\\
R3D-AD \cite{zhou2024r3d} & {69.6} & 73.6 & {90.0} & { 78.8} & 72.9 & { 75.2} & { 74.2} & { 63.0}    & { 75.7} & { 77.1} & { 72.1} & { 71.8} & { 74.9} &\\
ISMP \cite{ismp_2025} & 70.1 & 84.4 & 67.8 & 68.7 & 53.4 & 77.3 & 62.2 & 54.6 & 58.0 & 74.7 & 73.6 & 82.3 & 69.1 \\
PO3AD \cite{ye2025po3ad} & 57.3 & 74.5 & 68.1 & {85.8} &{ 74.2} & {95.2} & {82.1} & {67.5} & {85.2} & \textbf{96.6} & {73.9} & {83.0} & {83.9} & \\
\rowcolor{lightgray} Ours & \textbf{74.1} & \textbf{88.2} & \textbf{92.4} & \textbf{89.5} & \textbf{82.9} & \textbf{96.3} & \textbf{84.7} & \textbf{80.4} & \textbf{89.0} & {93.1} & \textbf{82.5} & \textbf{87.2} & \textbf{87.6} & {}\\
\bottomrule
\end{tabular}
}
\end{table*}


\begin{table}[!t]
\centering
\caption{Object-level AUC-ROC results (\%) of methods on the Real3D-AD dataset.}
\vspace{-0.8em}
\label{tab:cmps-real3dad}
\resizebox{0.9\linewidth}{!}{
\begin{tabular}{@{}l|ccccccc >{\columncolor{lightgray}}c}
\toprule
Category    
& \begin{tabular}[c]{@{}c@{}}CPMF\\ \cite{cao2024complementary} \end{tabular}       
& \begin{tabular}[c]{@{}c@{}}Reg3D-AD\\ \cite{liu2023real3d}\end{tabular}   
& \begin{tabular}[c]{@{}c@{}}IMRNet\\ \cite{Li_2024_CVPR}\end{tabular}        
& \begin{tabular}[c]{@{}c@{}}R3D-AD\\ \cite{zhou2024r3d}\end{tabular} 
& \begin{tabular}[c]{@{}c@{}}Group3AD\\ \cite{zhu2024towards}\end{tabular}        
& \begin{tabular}[c]{@{}c@{}}PO3AD\\ \cite{ye2025po3ad}\end{tabular}  
& \begin{tabular}[c]{@{}c@{}}ISMP\\ \cite{ismp_2025}\end{tabular}  
& \textbf{Ours} \\ 
\midrule
Airplane   & 70.1 & 71.6 & 76.2 & 77.2 & 74.4 & 80.4 & 85.8 & \textbf{87.5} \\
Car      & 55.1 & 69.7 & 71.1 & 69.3 & 72.8 & 65.4 & 73.1 & \textbf{75.6} \\
Candy     & 55.2 & 68.5 & 75.5 & 71.3 & 84.7 & 78.5 & 85.2 & \textbf{86.8} \\
Chicken  & 50.4 & \textbf{85.2} & 78.0 & 71.4 & 78.6 & 68.6 & 71.4 & 82.3 \\
Diamond & 52.3 & 90.0 & 90.5 & 68.5 & 93.2 & 80.1 & \textbf{94.8} & 90.1 \\
Duck  & 58.2 & 58.4 & 51.7 & 90.9 & 67.9 & 82.0 & 71.2 & \textbf{91.6} \\
Fish & 55.8 & 91.5 & 88.0 & 69.2 & \textbf{97.6} & 85.9 & 94.5 & 92.4 \\
Gemstone & 58.9 & 41.7 & 67.4 & 66.5 & 53.9 & 69.3 & 46.8 & \textbf{70.1} \\
Seahorse & 72.9 & 76.2 & 60.4 & 72.0 & 84.1 & 75.6 & 72.9 & \textbf{86.3} \\
Shell  & 65.3 & 58.3 & 66.5 & 84.0 & 58.5 & 80.0 & 62.3 & \textbf{83.7} \\
Starfish & 70.0 & 50.6 & 67.4 & 70.1 & 56.2 & 75.8 & 66.0 & \textbf{78.5} \\
Toffees & 39.0 & 82.7 & 77.4 & 70.3 & 79.6 & 77.1 & 84.2 & \textbf{85.9} \\ 
\midrule
Average  & 58.6 & 70.4 & 72.5 & 73.4 & 75.1 & 76.5 & 76.7 & \textbf{84.2} \\
\bottomrule
\end{tabular}}
\end{table}

\begin{table}[!th]
\centering
\caption{Point-level AUC-ROC results (\%) of methods on the Industry dataset.}
  \vspace{-1.0em}
\label{tab:industry-cmps}
\resizebox{\linewidth}{!}{
\begin{tabular}{@{}l|cccccccc|c@{}}
\toprule
Method 
& \begin{tabular}[c]{@{}c@{}}Bearing\\Cap\end{tabular}
& \begin{tabular}[c]{@{}c@{}}Dynamixel\\Plate\end{tabular}
& \begin{tabular}[c]{@{}c@{}}Franka\\Connector\end{tabular}
& Hull
& \begin{tabular}[c]{@{}c@{}}M4\\Tensioning\\Spool\end{tabular}
& \begin{tabular}[c]{@{}c@{}}Teeth\\Wrist\\Servo\end{tabular}
& \begin{tabular}[c]{@{}c@{}}Teeth\\Carpals\end{tabular}
& \begin{tabular}[c]{@{}c@{}}Wrist\\Pulley\\Base\end{tabular}
& Avg \\
\midrule

R3D-AD \cite{zhou2024r3d} & --- & --- & --- & 19.1 & 20.8 & --- & --- & --- & 20.0 \\
PO3AD \cite{ye2025po3ad} & 21.6 & 25.3 & 18.2 & 36.5 & 5.9 & 7.8 & 5.1 & 30.9 & 18.9 \\
ISMP \cite{ismp_2025} & 53.5 & 35.8 & 51.1 & 68.4 & 60.3 &  57.9 & 49.9 & 37.2 & 51.8 \\
\rowcolor{lightgray!20}
\rowcolor{lightgray} \textbf{Ours} & \textbf{67.8} & \textbf{64.3} & \textbf{75.6} & \textbf{78.1} & \textbf{82.3} & \textbf{85.9} & \textbf{89.5} & \textbf{61.2} & \textbf{75.6} \\
\bottomrule
\end{tabular}}
\end{table}

\subsection{Ablation study}
\paragraph{Time complexity \& memory} 
Our model achieves similar inference speed (178\,ms) to PO3AD (172\,ms) and R3DAD (183\,ms) while offering stronger performance on global anomalies. In terms of GPU memory usage, R3DAD requires 2318\,MB, PO3AD 2013\,MB, and ours 2125\,MB for the same point input. As shown in \cref{tab:inference_breakdown}, the 3D UNet backbone dominates computation (51.7\%), followed by the RoPE cross-attention decoder (29.4\%). FPS patchification accounts for 14.4\% of inference time, while the lightweight MLP head contributes minimal overhead (4.5\%). All timings were measured on a single NVIDIA RTX 3090 GPU with a batch size of 1.

\begin{figure}[h]
    \centering
    \includegraphics[width=0.75\linewidth]{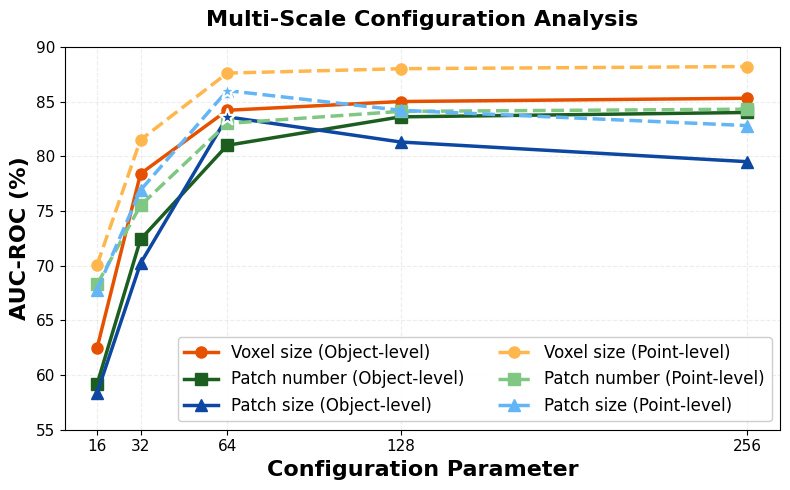}
    \vspace{-1.0em}
    \caption{AUC-ROC performance under varying configuration parameters. 
    Three parameter groups are compared: voxel size, patch number, and patch size. 
    Each parameter is evaluated at object-level (solid lines) and point-level (dashed lines).}
    \label{fig:parameter-sensitivity}
    \vspace{-0.4em}
\end{figure}

\begin{table}[!th]
\centering
\caption{Inference time breakdown of model components.}
\vspace{-0.8em}
\renewcommand{\arraystretch}{1.3}
\begin{adjustbox}{width=0.62\linewidth,center}
\begin{tabular}{lcc}
\toprule
\textbf{Module} & \textbf{Time (ms)} & \textbf{Percentage (\%)} \\
\midrule
3D UNet (MinkUNet34C) & 92.1 & 51.7 \\
RoPE Cross-Attention Decoder & 52.3 & 29.4 \\
MLP-based Head & 8.0 & 4.5 \\
FPS Patchification & 25.7 & 14.4 \\
\midrule
\textbf{Total} & \textbf{178.1} & \textbf{100.0} \\
\bottomrule
\end{tabular}
\end{adjustbox}
\label{tab:inference_breakdown}
\vspace{-0.6em}
\end{table}
\paragraph{Hyperparameter analysis}
\cref{fig:parameter-sensitivity} presents a sensitivity analysis of multi-scale patch settings on Real3D-AD~\cite{liu2023real3d}. Increasing the \textbf{voxel size} consistently improves both object-level and point-level AUC-ROC, with performance saturating beyond $128^3$, indicating sufficient geometric detail capture. The \textbf{patch number}, evaluated at voxel resolution $256^3$ and patch size $16$, yields steady gains before plateauing after $128$ patches, suggesting diminishing returns from finer spatial partitioning. In contrast, the \textbf{patch size}, tested at voxel resolution $256^3$ and patch number $32$, peaks at size $64$ and gradually declines at larger scales ($128$, $256$), as oversized patches blur fine-grained anomaly boundaries. Overall, moderate voxel resolution ($\geq 128^3$), sufficient patch numbers ($\geq 128$), and a balanced patch size ($64$) provide the best trade-off between local detail and global representation for robust anomaly detection and localization.

\paragraph{Patchification strategy} We evaluate the impact of different patchification strategies on performance, including semantic-part-based, FPS voxel-based, 3D grid, FPS sphere, and multi-scale sphere patchification. For semantic-part-based patchification, we use Find3D~\cite{ma20243d} with part-text prompts and PartField~\cite{partfield2025} on meshes, followed by point-cloud sampling with part labels. As shown, the multi-scale patchification achieves the best anomaly detection performance, while the semantic-part-based variant also attains comparable AUC-ROC to ISMP \cite{ismp_2025} in~\cref{tab:cmps-real3dad}. However, large semantic parts lack sufficient geometric granularity for fine feature matching. Additional details on patchification strategies are provided in the supplementary materials.

\begin{table}[!th]
\centering
\caption{Comparison of patch generation strategies (with figure example over column headers) in terms of object-level and point-level AUC-ROC performance (\%) on Real3D-AD dataset.}
\label{tab:patchify_comparison}
\vspace{-1.0em}
\begin{adjustbox}{width=0.86\columnwidth,center}
{\Large
\begin{tabular}{l|c|c|c|c|c}
\multirow{2}{*}{\centering \textbf{Metric}} 
& \multicolumn{5}{c}{
    \includegraphics[width=1.8\linewidth]{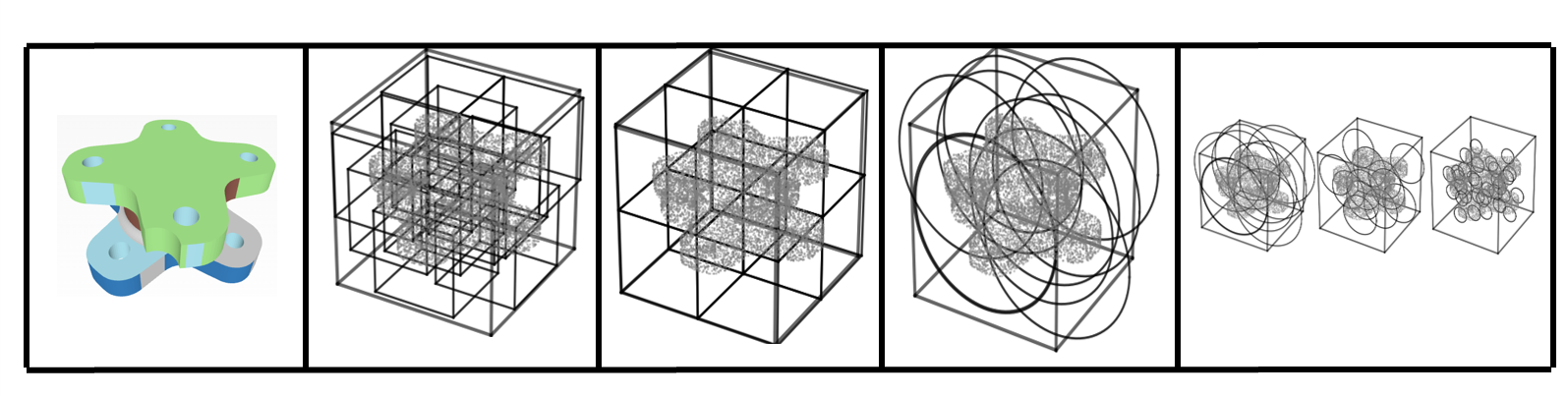}
} \\
& \textbf{Semantic Parts} & \textbf{FPS Voxels} & \textbf{3D Grids} & \textbf{FPS Spheres} & \textbf{Multi-scale Spheres} \\ \hline
\textbf{Object (\%)} & 75.1 & 81.4 & 80.3 & 81.9 & \textbf{84.2} \\ 
\textbf{Point (\%)} & 78.2 & 82.5 & 81.7 & 82.6 & \textbf{85.3} \\ \hline
\end{tabular}
}
\end{adjustbox}
\vspace{-1.4em}
\end{table}


\paragraph{Ablation of design choices.}
As shown in~\cref{tab:ablation-design}, removing negative augmentation significantly reduces the ability of the model to bridge the gap between normal and anomalous samples, resulting in degraded performance. We then compare three patch feature extraction methods: (1) direct pooling of point features $\in \mathbb{R}^{32}$, and (2) computing the mean point coordinates $\in \mathbb{R}^{3}$ within each patch and querying the pre-trained encoder for its feature, and (3) averaging all patch point features $\in \mathbb{R}^{32}$. Among these, the second approach, using mean point coordinates as queries, achieves the best performance. Finally, RoPE-based cross-attention proves essential for effective patch–point feature fusion: removing ML modulation of normal and anomalous features reduces accuracy by about 4\%, while disabling point-wise fusion leads to a significant drop of more than 10\%. Empirically, selecting the most relevant scale by max in \cref{eq:patch-similarity} outperforms fusing all scales (e.g., 84.2\% vs. 81.2\% AUC-ROC on Real3D-AD), since anomaly scales vary across categories and forced fusion can dilute discriminative features.

\begin{table}[!th]
\centering
\caption{Ablation study on the effectiveness of individual modules and pre-processing designs on the Real3D-AD~\cite{liu2023real3d} dataset (Best viewed when zoomed in.).}
\vspace{-0.8em}
\label{tab:ablation-design}
\begin{adjustbox}{width=1.0\linewidth,center}

\begin{tabular}{l|c|c|c|c|c}
\toprule
\textbf{Metric} 
& \textbf{No Neg-Aug} 
& \textbf{Pooling} 
& \textbf{Mean Point} 
& \textbf{Mean Feature} 
& \textbf{RoPE Cross-Atten} \\
\midrule
\textbf{AUC-ROC} & 68.2 & 77.8 & 79.4 & 76.5 & 84.2 \\
\textbf{AUC-PR}  & 65.1 & 75.9 & 73.6 & 78.3 & 82.7 \\
\midrule
\textbf{Metric} 
& \textbf{Vanilla Cross-Atten} 
& \textbf{No MLP Modul} 
& \textbf{No Cross-Atten} 
& \textbf{Multi-scale Modulation} 
& \textbf{Full} \\
\midrule
\textbf{AUC-ROC} & 79.1 & 80.3 & 72.6 & 81.2 & \textbf{84.2} \\
\textbf{AUC-PR}  & 77.4 & 78.5 & 70.8 & 79.3 & \textbf{82.7} \\
\bottomrule
\end{tabular}
\end{adjustbox}
\vspace{-1.5em}
\end{table}


\begin{figure}[th]
\centering
\includegraphics[width=0.46\columnwidth]{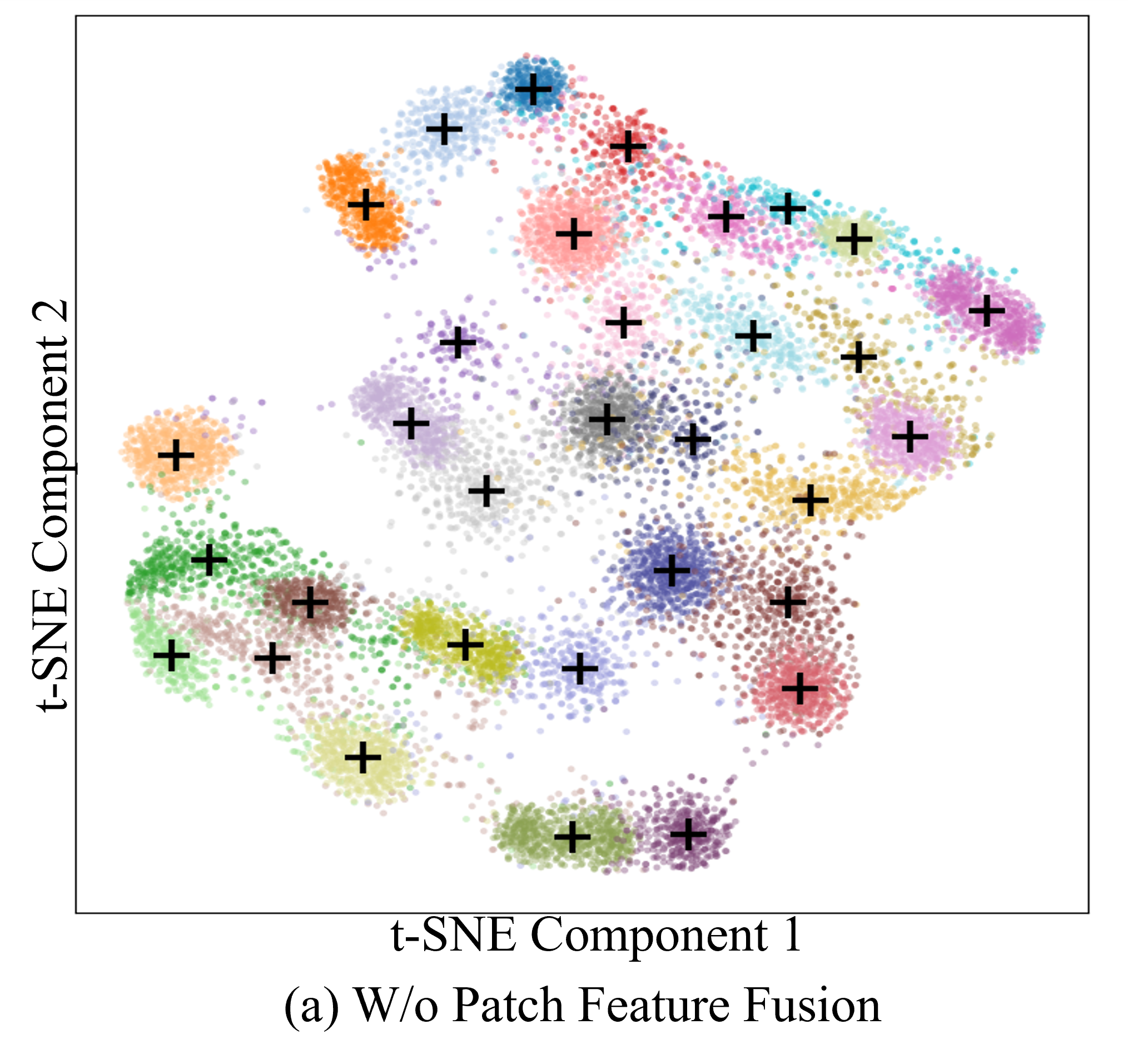}
\hfill
\includegraphics[width=0.46\columnwidth]{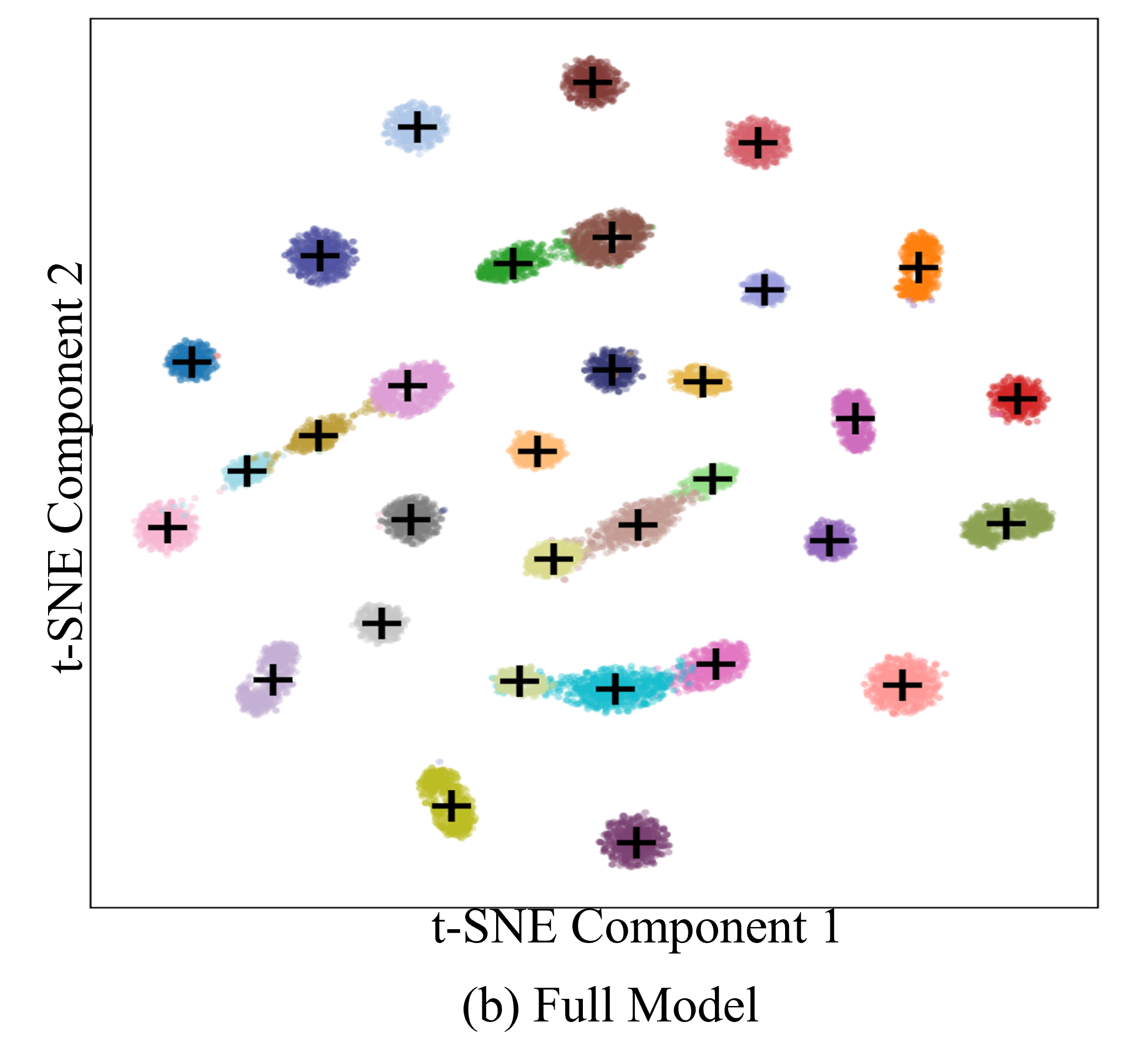}
\vspace{-1.0em}
\caption{t-SNE visualization of point feature distributions. `+' denotes patch centers; colors indicate 32 patch membership. (a) Without fusion, features scatter with poor separation, lacking discriminative structure despite close patch centers. (b) With patch feature modulation and fusion, clusters are well-separated.}
\label{fig:tsne_comparison}
\vspace{-0.5em}
\end{figure}
To better illustrate the benefits of patch features, we present the results in~\cref {fig:tsne_comparison} for 32 patches on the ``starfish'' class. For the tests, the model is trained for an equivalent of 1000 epochs. Without fusion (\cref{fig:tsne_comparison}a), point features scatter with poor separation, indicating weak discriminative power for reliable anomaly localization. With our patch feature modulation and fusion, the features form well-separated clusters with clear margins (\cref{fig:tsne_comparison}b), enabling more accurate anomaly discrimination for regression. For additional ablation results on patch parameters such as patch size and patch number, please refer to the supplementary material.

\section{Conclusion}
\label{sec:conclusion}
We propose a multi-scale patch-based framework for shape anomaly detection. We further introduce a real industrial test set with planar and angular displacement defects.
\vspace{-1.0em}
\paragraph{Limitations \& future work.} Our method remains sensitive to sampling randomness. Future work will explore incorporating topological or semantic part cues to produce more consistent patches and scale to real industrial data.


\section*{Acknowledgements}
The first two authors acknowledge the financial support from the University of Melbourne through the Melbourne Research Scholarship. This research was supported by the University of Melbourne's Research Computing Services and the Petascale Campus initiative.

{
    \small
    \bibliographystyle{ieeenat_fullname}
    \bibliography{main}
}


\end{document}